\documentclass[conference]{IEEEtran}

\IEEEoverridecommandlockouts
\usepackage{cite}
\usepackage{amsmath,amssymb,amsfonts}
\usepackage{algorithmic}
\usepackage{graphicx}
\usepackage{textcomp}
\usepackage{xcolor}
\usepackage{booktabs,siunitx}
\usepackage[caption=false]{subfig}
\usepackage{booktabs,caption}
\usepackage{url}
\usepackage[flushleft]{threeparttable}

\makeatletter
\def\ps@IEEEtitlepagestyle{%
  \def\@oddfoot{\mycopyrightnotice}%
  \def\@evenfoot{}%
}
\def\mycopyrightnotice{%
  {\footnotesize Preprint of accepted paper in IEEE International Conference on Fuzzy Systems 2019 \hfill}
  \gdef\mycopyrightnotice{}
}

\def\BibTeX{{\rm B\kern-.05em{\sc i\kern-.025em b}\kern-.08em
    T\kern-.1667em\lower.7ex\hbox{E}\kern-.125emX}}

\begin{document}

\title{Fuzzy Rough Set Feature Selection to Enhance
Phishing Attack Detection}

\author{
	\IEEEauthorblockN{Mahdieh Zabihimayvan and Derek Doran}
	\IEEEauthorblockA{
			   Department of Computer Science and Engineering
			\\ Wright State University, Dayton, OH, USA 
			\\ \{zabihimayvan.2, derek.doran\}@wright.edu
	}
}

\maketitle

\begin{abstract}
Phishing as one of the most well-known cybercrime activities is a deception of online users to steal their personal or confidential information by impersonating a legitimate website. Several machine learning-based strategies have been proposed to detect phishing websites. These techniques are dependent on the features extracted from the website samples. However, few studies have actually considered efficient feature selection for detecting phishing attacks.
In this work, we investigate an agreement on the definitive features which should be used in phishing detection. We apply Fuzzy Rough Set (FRS) theory as a tool to select most effective features from three benchmarked data sets. The selected features are fed into three often used classifiers for phishing detection. To evaluate the FRS feature selection in developing a generalizable phishing detection, the classifiers are trained by a separate out-of-sample data set of 14,000 website samples. The maximum F-measure gained by FRS feature selection is 95\% using Random Forest classification. Also, there are 9 universal features selected by FRS over all the three data sets. The F-measure value using this universal feature set is approximately 93\% which is a comparable result in contrast to the FRS performance. Since the universal feature set contains no features from third-part services, this finding implies that with no inquiry from external sources, we can gain a faster phishing detection which is also robust toward zero-day attacks.

\end{abstract}


\section{Introduction}
Phishing is categorized as a social engineering attack to coheres users to perform adverse actions on behalf of an attacker~\cite{basnet2008detection}. 
A usual purpose of such attacks is to deceive users to steal their personal or confidential information. To this end, phishers try to mimic the emails or web pages that indicate a high visual similarity to the legitimate emails or web pages. The pages are further accompanied with at least one login form to gather personal or confidential 
user information~\cite{abu2007comparison}. The content of phishing emails and web pages are 
developed from social engineering practices so that victims are persuaded to follow the instructions with the hope of, for example, updating or validating their information, finding a job, winning a big prize, or receiving a significant discount to a service. 
According to the most up-to-date trend reports by the Anti-Phishing Working Group~\cite{APWG}, over 151,014 unique phishing websites and 
270,557 unique emails have been documented. These sites and emails
mimicked at least 777 brands targeted by phishing campaigns during the third quarter of 2018.

A range of technical strategies have been proposed to detect phishing emails and websites. The strategies can broadly be categorized into three classes:~(i) list-based;~(ii) heuristic; and~(iii) and machine learning (ML)-based approaches~\cite{rami2014}. The first class simply
references different lists of legitimate and phishing URLs. While these
lists are frequently updated, there is an inevitable time gap between 
the emergence of a phishing website, its observation, reporting, and 
finally, addition to a suspicious URL list. This makes list-based
strategies vulnerable towards emerging and zero-day phishing 
attacks~\cite{jain}. Heuristic approaches can automatically 
recognize a phishing web pages based a suite of features extracted from them~\cite{heuristic}. One method, for example, focuses on the similarity of the suspicious web page with the legitimate one based on visual features such as image, logo, and textual content of the page~\cite{visual}. However, there is no guarantee that a phishing site and 
a legitimate web page or email has features in common, which can cause poor detection accuracy in practice. Furthermore, heuristic techniques can be bypassed 
by a knowledgeable attacker that ensures their phishing website does not contain the detected features. ML methods~\cite{mlBased} are also dependent on the set of features extracted for each web page and further require ground truth phishing and legitimate websites for training. The quality and variety of 
websites in this training set has a strong effect on the final detection accuracy~\cite{rao2018detection}, and it can be expensive to obtain a training set that is suitably sized and diverse.

Despite these challenges, ML approaches for phishing detection have been 
an active area of research. Several studies have been conducted on different data sets using various classification algorithms~\cite{Feng2018,mohammad2013predicting,sahingoz2019machine}. There is an undeniable effect 
of the features used for classification on the accuracy of an 
algorithm~\cite{rajab2018anti}, but few studies have actually considered 
intelligent methods feature selection. The problem of feature selection is 
crucial to build phishing detection systems that are generalizable in practice. 
For example, it is feasible that the set of features best distinguishing 
phishing websites that imitate a financial institution may be different from
sites imitating an e-commerce platform, or of websites that are encoded in
different languages. The present art in feature selection are simply
based on heuristic thresholds~\cite{rajab2018anti,babagoli2018heuristic}. 

To address this problem, we apply Fuzzy Rough Set (FRS) theory~\cite{pawlak1982rough} as a tool to select the most effective features for detecting phishing websites in this paper. Our feature selection scheme
preprocesses features that are fed into three often used classification methods
for phishing detection: a multilayer perceptron, a random forest, and SMO. 
We train each classifier on a separate out-of-sample dataset to examine the ability of FRS feature selection in developing a generalizable 
phishing detection. This training set which is composed by random extraction of websites from an online repository~\cite{chiew2018building} contains 7,000 legitimate and 7,000 phishing website samples. 
The benefit of using FRS for 
feature selection is realized by comparing our results
against three recently proposed algorithms~\cite{rajab2018anti,babagoli2018heuristic}.

The rest of this paper is organized as follows: Section~\ref{sec:relatedwrok} discusses the related work on phishing detection. Section~\ref{sec:featureS} presents the procedure used for feature selection. Section~\ref{sec:experiment} evaluates the proposed method and compares its performance with other related algorithms. Finally, Section~\ref{sec:conclusion} summarizes the main conclusions. 

\section{Related Work}
\label{sec:relatedwrok}

ML-based approaches to detect phishing websites is an active research area that  employs a wide range of supervised classification techniques to segregate phishing class. Feng {\em et al.} propose a novel neural network for phishing detection~\cite{Feng2018}. They improve the generalization ability of the network by designing risk minimization principle. Performance of the proposed network is evaluated over a UCI repository~\footnote{\url{https://archive.ics.uci.edu/ml/datasets/phishing+websites}} containing 11,055 samples labeled as phishing/legitimate. The dataset also specifies 30 features for each website categorized as Address bar-based, Abnormal-based, HTML/Javascript-based, and Domain-based features. Rao and Pais propose a novel algorithm to detect phishing websites using both machine learning techniques and image checking. They also extract features from URL, website content, and third-party services~\cite{rao2018detection}. It is worth mentioning that although using the features from third-party service can increase the detection time, it increases the detection accuracy in practice~\cite{sahingoz2019machine}. They evaluate the performance of the proposed algorithm over 1407 legitimate and 2119 phishing websites collected from PhishTank~\footnote{\url{http://www.phishtank.com/}} and Alexa database~\footnote{\url{http://www.alexa.com/}}, respectively. 

Mohammad {\em et al.} propose a novel self-structuring neural network for detecting phishing websites~\cite{rami2014}. They specify 17 features, some extracted from third-party service, for 600 legitimate and 800 phishing websites collected from PhishTank and Millersmiles~\footnote{\url{http://www.millersmiles.co.uk/}} archives. Their experiments indicate the high generalizability and ability of the neural network in phishing detection. In another work, they propose a feed forward neural network trained by back propagation to classify websites~\cite{mohammad2013predicting}. 18 features are specified for 859 legitimate and 969 phishing websites, respectively. Jain and Gupta propose a machine learning-based technique using only client-side features to detect phishing websites~\cite{jain2018towards}. They extract 19 features from the URL and source code of the web pages and evaluate their method over 2,141 phishing web pages from PhishTank and Openfish~\footnote{\url{https://openphish.com/}} and 1,918 legitimate pages from Alexa database, some online payment, and banking websites.

Although all the above-mentioned studies have suggested different features to detect phishing websites, some features may not be sufficiently discerning to highlight phishing instances~\cite{rajab2018anti}. Only a limited number of work has focused on selecting the most effective features to detect phishing websites. Rajab proposes using Correlation Feature Set (CFS) and Information Gain (IG) to select the most influential features to detect phishing activities~\cite{rajab2018anti}. The results over the UCI repository containing 30 features specified for 11,055 samples indicate 11 and 9 features selected by IG and CFS, respectively. The classification performance using selected features is also evaluated by a data mining method called RIPPER. Similarly, Babagoli {\em et al.}use a similar data set and propose feature selection using decision trees and the
wrapper method~\cite{rodrigues2014wrapper} which results in selecting 20 features~\cite{babagoli2018heuristic}. They evaluate the phishing detection performance using a novel meta-heuristic-based nonlinear regression algorithm. 
Still, the feature selection methods proposed by these studies are dependent on the data and requires user-specified threshold values which should be set heuristically. These thresholds can affect the ultimate performance of the 
classification algorithm, especially when features are selected based on 
out-of-sample training data in practice. 

In this work, we propose using FRS theory as the feature selection algorithm.
In contrast to the related work focusing on only one data set, our experiments 
evaluate the generalizability of our approach by finding a universal set of 
discriminative features from   
three benchmark data sets. The datasets include a new source
of 7,000 legitimate and 7,000 phishing website samples randomly extracted from the a phishing website repository~\cite{chiew2018building}. We compare the detection performance using the selected features with the results of features selections proposed in~\cite{rajab2018anti}~\cite{babagoli2018heuristic}.

\section{Fuzzy Rough Set Feature Selection}
\label{sec:featureS}

Rough Set (RS) theory, developed by Pawlak~\cite{pawlak1982rough}, is a method to decide how to separate a set of data, with a decision boundary and an indiscernibility relation $R$. Supposing each set member $s_i$ described by $D$ features with discrete values, $R$ defines a binary 
similarity between two members
$s_m$ and $s_n$ as follows:
$$
R(s_m,s_n)= 
\begin{cases}
    1 & \text{if } \forall \; i \in \{1,...,D\} \;\; s_m(i)=s_n(i)\\
    0 & \text{if } \exists \; i \in \{1,...,D\} \text{ where } s_m(i) \neq s_n(i)\\
\end{cases}
$$
Similar samples join individually together in one exclusive class, $[s_i]_R$, which is defined as follows:
$$
[s_i]_{R}=\{s_j \;|\; R(s_i,s_j)=1\}
$$
\begin{figure}
\includegraphics[width=0.45\textwidth]{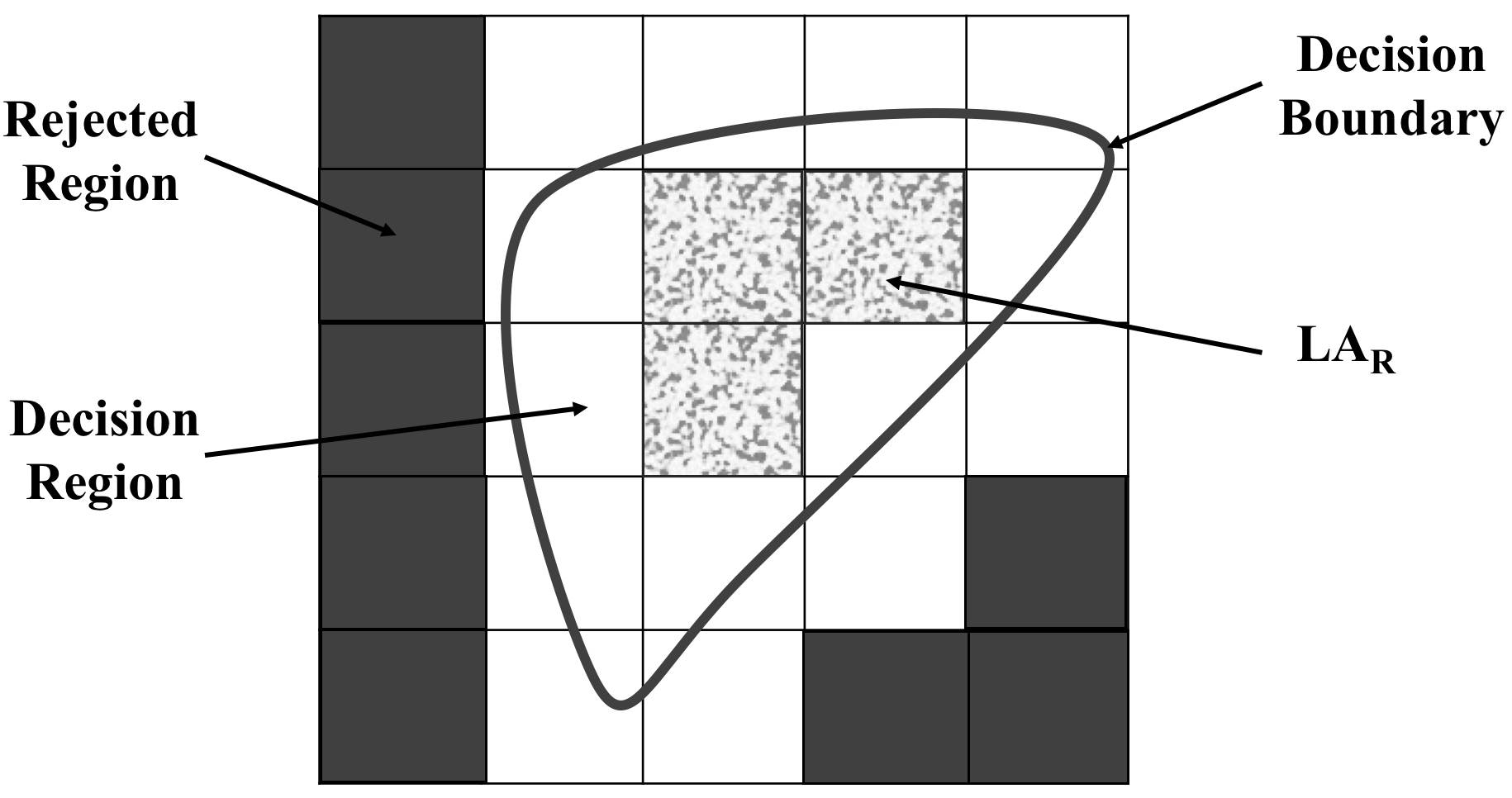}
\caption{The universe of discourse regions segregated by RS}
\label{fig:frs}
\vspace{-8px}
\end{figure}
To better understand the definition of $[s_i]_R$, figure~\ref{fig:frs} illustrates
a universe of discourse as a grid of equivalence class (squares) representing
set members that have identical feature values ($R=1$). The closed curved line  
further defines a hypothetical decision boundary. The dotted squares show regions that are fully located within the decision boundary. Such regions are equivalence classes are fully describable by the boundary and are called the boundary’s lower approximation set $LA_{R}$. The boundary region also contains the sets of equivalence classes (white squares) that are both inside and outside of the decision boundary. This boundary region unioned with $LA_{R}$ forms the boundary’s upper approximation set $UA_{R}$. Finally, the dark squares lying outside of the decision boundary specify the samples that are rejected by the decision boundary. Supposing $f$ is the function defining the decision boundary and $U$ indicating the universe of discourse, $UA_{R}$ and $LA_{R}$ are declared as:
$$
LA_{R}= \{ [s_i] \in U \;|\; \forall s_j \in [s_i]_R: f(s_j) \leq 0 \}
$$
$$
UA_{R}= \{ [s_i] \in U \;|\; \exists s_j \in [s_i]_R: f(s_j) \leq 0 \}
$$
For features with continuous values, RS is extended by Fuzzy Rough Set (FRS)~\cite{dubois1990rough} theory. FRS asserts that there is a subset of features representing
the {\em labels} of a set member 
while the remaining features are non-label. Assuming the feature values are normalized between 0 and 1, the similarity 
relation between two set members $R$ is given by: 
$$
\forall  i \in F,L: R(s_m,s_n)=\tau(\{max(0,1-\Delta_2(s_m(i),s_n(i)))\})
$$
where $F$ and $L$ are the sets of non-label and label features, $s_m(i)$ shows the value of $i$th feature of sample $s_m$, and $\Delta_2(x,y)$ defines the squared difference between $x$ and $y$.
$\tau$ is the Lukasiewicz t-norm~\cite{radzikowska2002comparative} that aggregates the similarity of all the features of the two samples and returns a values in $[0,1]$. The equation below indicates how to calculate the $\tau$ where $x$ and $y$ are both in $[0,1]$:
$$
\tau(x,y)=max(0,x+y-1)
$$
To determine if a set member $s_m$ is in the upper or lower approximation set of a decision boundary, FRS defines 
Upper and Lower Approximation Memberships $\mu_U$ and $\mu_L$ as:
$$
\mu_U(s_m)=\sup_{s_n\in U, s_n\neq s_m} \tau(R_F(s_m,s_n),R_L(s_m,s_n))
$$
$$
\mu_L(s_m)=\inf_{s_n\in U, s_n\neq s_m} I(R_F(s_m,s_n),R_L(s_m,s_n))
$$
where $R_F(s_m,s_n)$ shows the similarity between $s_m$ and $s_n$ using only non-label features while $R_L(s_m,s_n)$ gives the similarity using only label features set. $I$ is the fuzzy Implicator operator that is declared as follows where again 
$q,s \in [0,1]$:
$$
I(q,s)=min(1,1-q+s)
$$
A higher value of $\mu_L(s_m)$ gives us more confidence that $s_m$ is fully describable by the decision boundary. The values greater than zero for this membership function are considered for feature selection used in this work.
\section{Experiments}
\label{sec:experiment}

\begin{table}
	\centering
\caption{Data set description}
\begin{tabular}{|c|c|c|c|c|}
\toprule
{Dataset} & Year & {Size} & Num. Features & {Num. Phishing Sites} \\
\midrule
UCI1 & 2014 & 11,055 & 30 & 3,793 \\
Mendeley & 2018 & 10,000 & 48 & 5,000 \\
UCI2 & 2014 &1,353 & 9 & 702 \\
\bottomrule
\end{tabular}
\label{tab:datasets}
\end{table}

We apply FRS to select features from three benchmark data sets for phishing attacks. Table~\ref{tab:datasets} indicates the description of data used in this work. The first data set is from the UCI repository, UCI1\footnote{\url{https://archive.ics.uci.edu/ml/datasets/phishing+websites}}, that contains 30 features over 11,055 samples labeled as phishing/legitimate websites. Based on~\cite{rami2014}, the proposed features of this data set are categorized into 4 groups: Address bar-based, Abnormal-based, HTML/Javascript-based, and Domain-based features. Table~\ref{tab:featureClasses} presents a description of these classes in more detail. The second data set, Mendeley~\cite{mendeley}, specifies 48 features for 5,000 legitimate and 5,000 phishing websites. According to the feature class descriptions in Table~\ref{tab:featureClasses}, these 48 features can be also categorized into three groups of Address bar-based, Abnormal-based, and HTML/Javascript-based features. Also, Table~\ref{tab:d1d2} represents 23 features of this data set which are shared with the features in UCI1. The UCI2 dataset\footnote{\url{https://archive.ics.uci.edu/ml/datasets/Website+Phishing}} specifies 9 features from across the four feature classes over 
702 phishing, 548 legitimate, and 103 suspicious websites. Suspicious website has features of both legitimate and phishing samples and thus, can be considered as either phishy or legitimate~\cite{abdelhamid2014phishing}.

\begin{table}
	\centering
\caption{Feature classes across the three data sets}
\setlength{\tabcolsep}{3pt} 
\renewcommand{\arraystretch}{1.5} 
\begin{tabular}{ll}
\toprule
Feature Class & Description\\
\midrule
Address bar-based &\parbox{5.5cm}{Features of the web page URL such as its length in character, port, and the protocol used to transfer information.}\\[8pt]
Abnormal-based&\parbox{5.5cm}{Features of abnormal activities on a page, 
such as loading embedded objects from external domains or submissions to an email address.}\\[8pt]
HTML/Javascript-based&\parbox{5.5cm}{Features of HTML and Javascript functions embedded in the source code of the page.}\\
Domain-based&\parbox{5.5cm}{Features related to information from third-party services.}\\
\bottomrule
\end{tabular}
\label{tab:featureClasses}
\vspace{-5px}
\end{table}

The differences in sample size and number of features among the three data sets indicate some differences in the characteristics of the data studied in this work. In UCI1, the number of legitimate websites is more than twice of the number of phishing samples while in Mendeley, these classes have equal population size. In UCI2, although the sample sizes of both classes are approximately similar and have the dominant population, there is a minority class for suspicious samples. Furthermore, in contrast to the UCI repositories published in 2014, the Mendeley data has been published recently and thus, can be a representative of more updated samples for phishing websites. Finally, in contrast to some related work focusing on small number of data~\cite{mohammad2013predicting,rami2014,rao2018detection,peng2018detecting}, the sample size of data sets we consider is 
notably high. 

\begin{table}
\begin{threeparttable}
\caption{Features shared between UCI1 and Mendeley}
\vspace{-5px}
\setlength{\tabcolsep}{3.8pt} 
\renewcommand{\arraystretch}{1.5} 
\begin{tabular}{llll}
\toprule
No.&Feature Name&Type&Description \\
\midrule
1&UrlLen& D&length of URLs. \\
2&ExtFavicon&B&\parbox{3.5cm}{Favicon is from a different domain.}\\[8pt]
3&NumDash& D&No. dash in webpage URL.\\
4&NumDashInHostname&D& \parbox{3.5cm}{No. dash in URL host name.}\\[2pt]
5&AtSymbol&B&@ exists in the URL.\\
6&NoHttps&B&HTTP exists in the URL.\\
7&IpAddress&B&URL host name contains IP.\\
8&DomainInPaths&B& \parbox{3.5cm}{URL path contains TLD/ccTLD.}\\[8pt]
9&UrlLenRT&C&\parbox{3.5cm}{No. all characters in the webpage URL.}\\[8pt]
10&PctExtResUrls&C&\parbox{3.5cm}{\% external resource URLs in the HTML source code.}\\[8pt]
11&PctExtHlinks&C&\parbox{3.5cm}{\% external links in the HTML source code.}\\[8pt]
12&ExtFormAct&B&\parbox{3.5cm}{An external domain URL in the form action attribute.}\\[8pt]
13&AbnormFormAct&C&\parbox{3.5cm}{Occurring abnormal strings (\#, about:blank, empty string, javascript:true) in form action.}\\[10pt]
14&PctNullSelfRedirHlinks&C&\parbox{3.5cm}{\% hyperlink fields of current URL, empty, abnormal, or self-redirect values.}\\[8pt]
15&Submit2Email&B&\parbox{3.5cm}{HTML source code contains “mailto”.}\\[8pt]
16&PctExtResUrlsRT&C&\parbox{3.5cm}{\% external resource URLs in webpage HTML source code.}\\[8pt]
17&AbnormExtFormActR&C&\parbox{3.5cm}{Occurring empty string, foreign domain, about:blank in form action.}\\[8pt]
18&ExtMetaScriptLinkRT&C&\parbox{3.5cm}{\% meta, script and link tags containing external URLs.}\\[8pt]
19&PctExtNullSelfRedirHlinksRT&C&\parbox{3.5cm}{\% hyperlinks in HTML source code with different domain names.}\\[8pt]
20&RightClickDisabled&B&\parbox{3.5cm}{HTML source code contains command to disable right click.}\\[8pt]
21&PopUpWin&B&\parbox{3.5cm}{HTML source code contains pop-ups.}\\[8pt]
22&FakeLinkInStatusBar&B&\parbox{3.5cm}{HTML source code contains onMouseOver.}\\[8pt]
23&IframeOrFrame&B&\parbox{3.5cm}{HTML source code contains frame/iframe.}\\[8pt]
\bottomrule
\end{tabular}
\begin{tablenotes}
      \small
      \item The column Type indicates the value types of features. D, B, and C show discrete, binary, and categorical values, respectively.
\end{tablenotes}
\label{tab:d1d2}
\vspace{-12px}
\end{threeparttable}
\end{table}

To evaluate the detection performance using features selected by FRS, we employ three well-known classification algorithms: the multilayer-perceptron, random forest, and sequential minimal optimization (SMO) that have been extensively employed for phishing detection by other related studies~\cite{jain2018towards,rajab2018anti,sahingoz2019machine,rao2018detection}. All hyper-parameters are also set based on the values used in these previous works. 
To evaluate the ability for FRS feature selection to develop a generalizable 
phishing detector, we train each classifier on a separate out-of-sample dataset
composed by random extraction of phishing and non-phishing websites from an
online repository~\cite{chiew2018building}. This training set has 7,000 legitimate and 7,000 phishing website samples. 

We compare the results using 
FRS against three recently proposed feature selection algorithms: information gain (IG),
correlated feature set~\cite{rajab2018anti} (CFS), and using a combination of decision tree and 
the Wrapper method~\cite{babagoli2018heuristic} (DW).
IG chooses features from a ranking based on the level of information each provides with respect to the target class (features which specifies the class labels)~\cite{rajab2018anti}. CFS sorts features based on their correlation with the target class. The first-ranked feature has highest correlation with the target class and lowest correlation with the others~\cite{rajab2018anti}. DW utilizes a combination of decision tree and Wrapper to select an optimal feature set~\cite{babagoli2018heuristic}. Since IG and CFS just provide a ranking of features, a threshold is needed to specify which features should be selected as efficient. Furthermore, DW is dependent on the expert opinion to decide what decreasing in the accuracy should be considered as significant to stop the method. In this work, the thresholds and other settings are set based on the values suggested in~\cite{rajab2018anti} and~\cite{babagoli2018heuristic}.
\begin{figure}
\subfloat[UCI1]{\includegraphics[width = 3.2in]{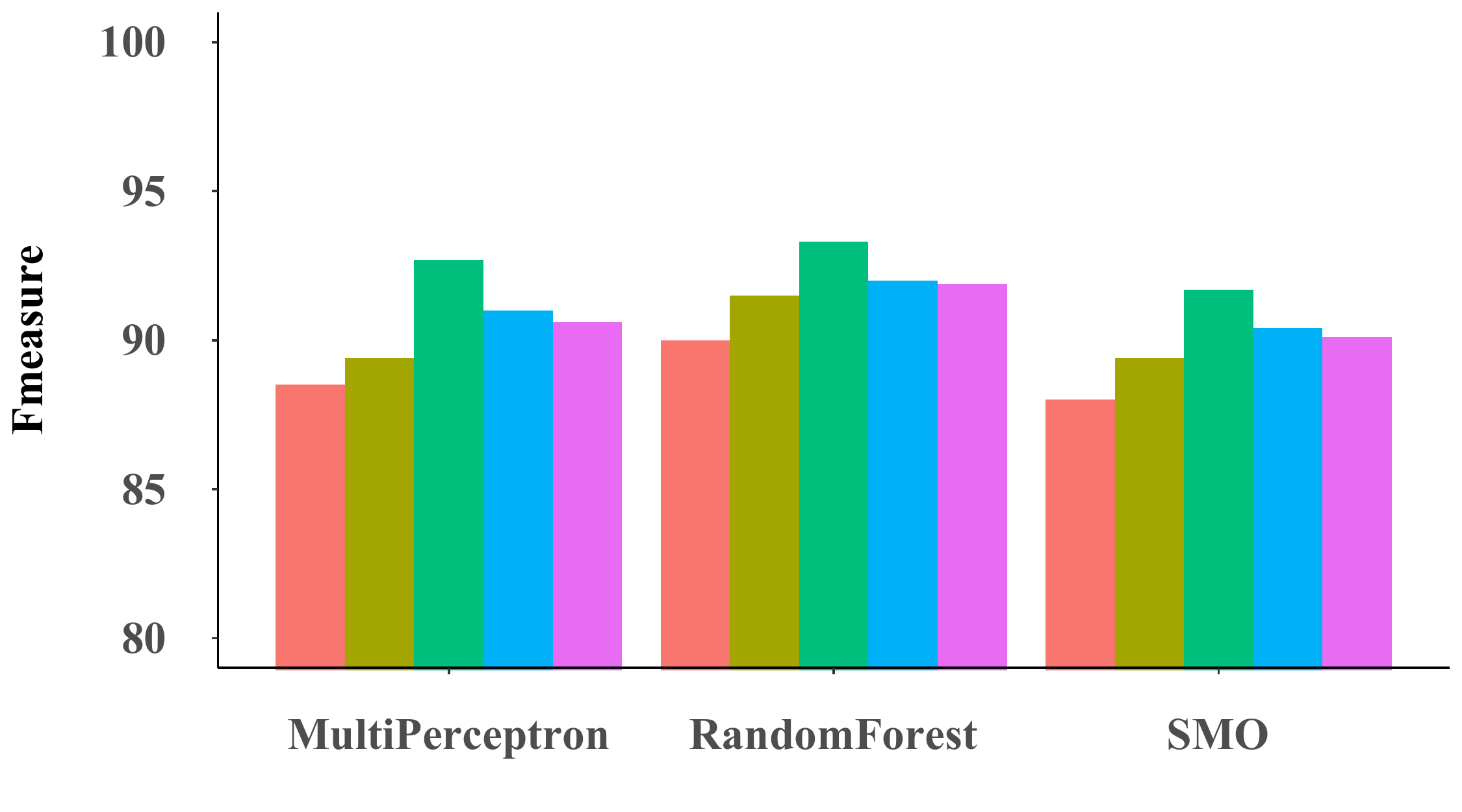}} \\
\subfloat[Mendeley]{\includegraphics[width = 3.2in]{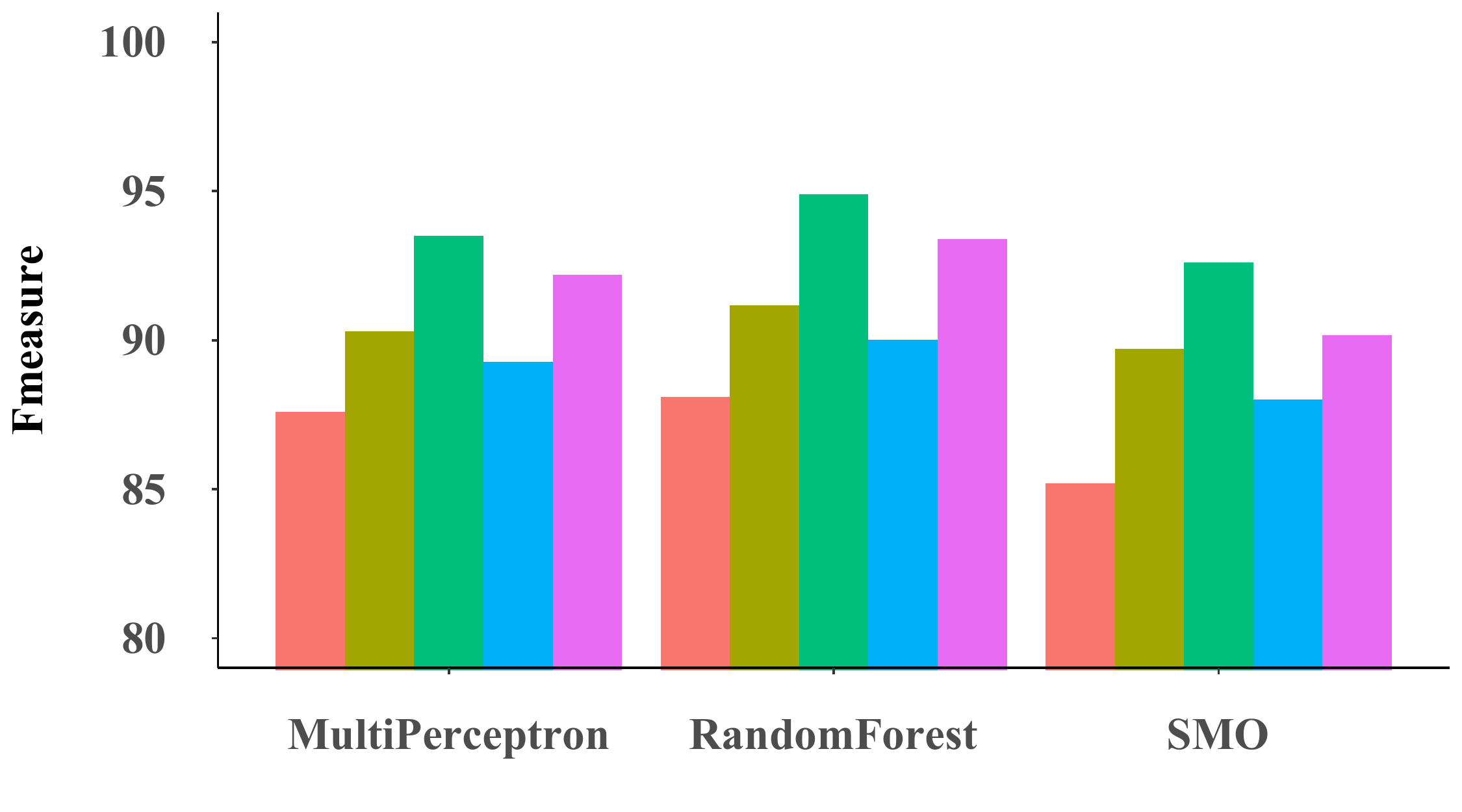}} \\
\subfloat[UCI2]{\includegraphics[width = 3.2in]{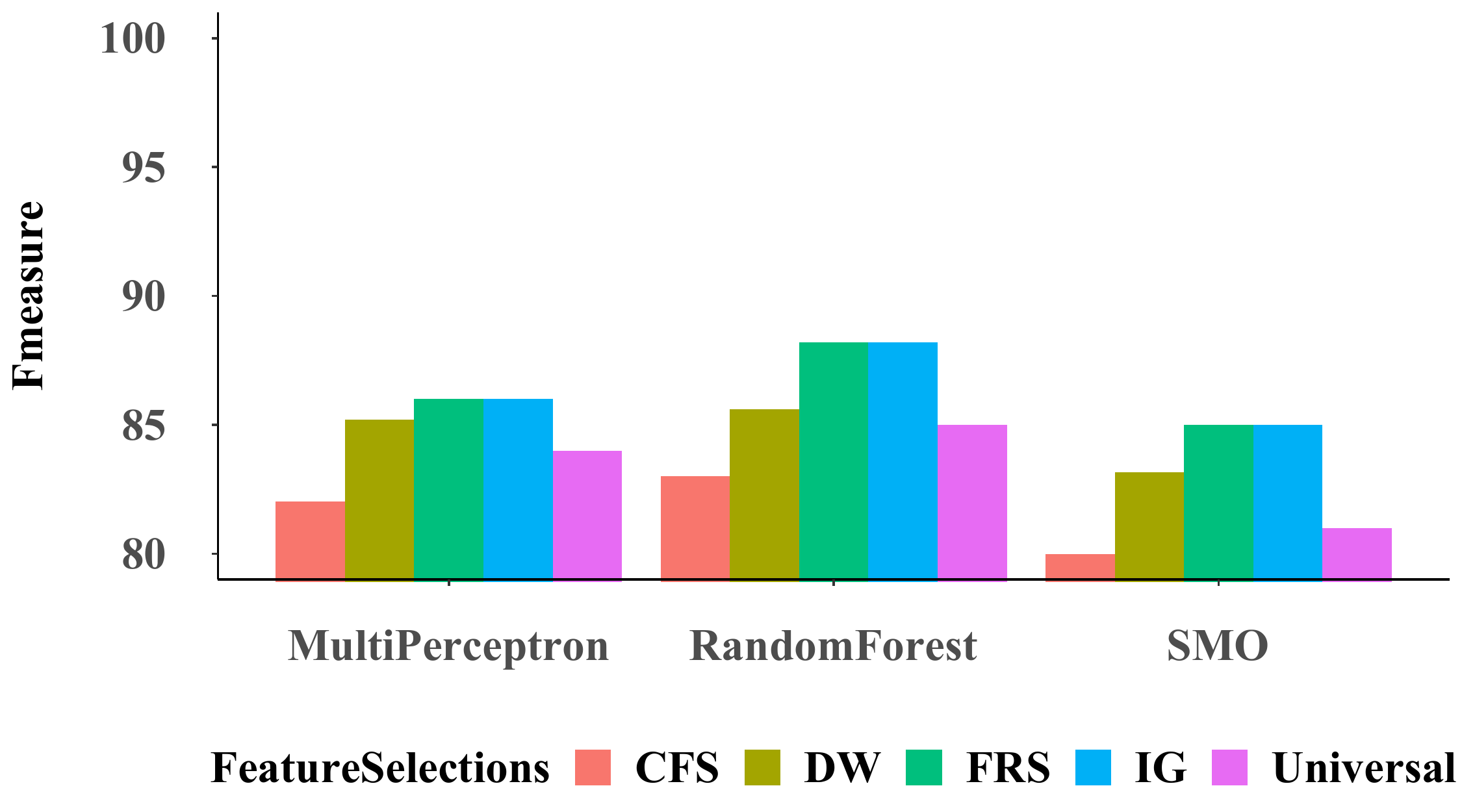}}\\
\caption{F-measure for different classifiers and features sets}
\label{fig:fvals}
\vspace{-5px}
\end{figure}

Figure~\ref{fig:fvals} presents the F-measure values for classifiers using four feature selection algorithms over each data set. Assuming phishing and legitimate samples belong to positive and negative classes respectively, F-measure calculates the quality of classification for both classes~\cite{mprediction2018}. This metric is defined as $2\times \frac{Precision\times Recall}{Precision+Recall}$ where $Precision=\frac{TP}{TP+FP}$ indicates number of truly detected phishing samples over number of all samples identified as phishing. Also, $Recall=\frac{TP}{TP+FN}$ is the number of truly identified phishing samples over the number of all existed phishing websites. 

Across all datasets, the random forest model 
yields the best performance no matter the feature selection method considered. 
Over UCI1 and Mendeley, the FRS performance is higher than the other feature selection algorithms. The identical performance of 
FRS and IG over UCI2 is also the result of both methods selecting all features, 
which could be a byproduct of the small size of this dataset. 
The small number of features on UCI2 also contribute to the comparatively 
low performance of all classifiers on this dataset. 
In contrast to IG, CFS, and DW which are dependent on some thresholds set heuristically, FRS discovers the data dependencies for feature selection based on the data with no need for additional information. In other words, FRS enables handling incomplete information to select effective features based on fuzzy set theory that investigates vagueness in rough set framework~\cite{sadeghi2018automatic}.  

We further investigate the types of features selected by each 
method. 
Table~\ref{tab:features} presents the list of features selected by FRS for each data set. 
24, 30, and 9 features have been selected for UCI1, Mendeley, and UCI2, respectively. As indicated, 
7 features selected for UCI1 are Domain-based which utilize third-party services such as Alexa or WHOIS databases 
to get more information about phishing samples. However, Mendeley does not use any Domain-based feature and as 
Figure~\ref{fig:fvals} indicates, 
the maximum performance over Mendeley is approximately 95\% which outperforms all the FRS results over UCI1. It shows using domain-based features not only increases the detection time due to inquiry from external sources~\cite{sahingoz2019machine}, but also has no significant effect to improve the detection performance.

The bold features in Table~\ref{tab:features} are shared features among all three data sets that have been selected by FRS. FRS identifies the four Address Bar-based features UrlLen, PrefSuff, HaveSubDomain, and Favicon, accompanied with the five Abnormal-based features of ReqUrl, UrlAnchor, LinksInTags, SFH, and Submit2Email to be ``universally applicable'' features
for phishing detection across all datasets. As phishers often use long URLs, UrlLen is a discrete feature to consider the length of web page URL. Dash in URLs is also another signal for phishing websites which is considered by PrefSuff that checks the existence of prefixes or suffixes separated by dash in the URL. HaveSubDomain examines another way of confusing the website visitors which modifies the number of subdomains in the URL address. Favicon also checks if the favicon indicated in the address bar is loaded from another domain. ReqUrl is an Abnormal-based feature which considers if the external objects within the webpage are loaded from another domain. UrlAnchor checks whether an anchor within the web page has different domain or has no link to other pages. Legitimate websites often use meta, script, and link tags in their HTML document. LinksInTags examines the percentage of such tags in the web page source code. The value of Server Form Handler (SFH) can be another way to distinguish phishing pages. The empty string or ``about:blank'' in the SFH can indicate legitimate website while different domain name in SFH shows a phishing effort. Finally, using a client-side function called ``mailto:'' to submit user information to an external server can indicate a phishing attack. This examines by the Boolean feature called Submit2Email.

As mentioned above, the universal features are based on the general signals in the address bar or the HTML document of the web page regardless of its application (e-commerce, online banking, etc), its dominant language, or its statistical reports in external sources like Alexa. The fifth bar for each data set in Figure~\ref{fig:fvals} indicates the performance using only these universal features. Although on UCI2, the low number of features belonging to the universal feature set (only 4) reduces the detection performance, the maximum performance of universal features ($\approx 93\%$) competes with the maximum FRS performance ($\approx 95\%$) over UCI1 and Mendeley. It indicates that although the universal feature set has lower number of features, its F-measure value is comparable with the result of the larger set of features selected by FRS. Since no domain-based feature belongs to the universal feature set, this finding can not only reduce the detection time and make it robust towards zero-day attacks, it can solve the curse of dimensionality for classification techniques~\cite{2017soft}~\cite{hamidzadeh2018detection}.
 
\begin{table}
\begin{threeparttable}
\caption{Features selected by FRS over the data sets}
\setlength{\tabcolsep}{-1pt} 
\renewcommand{\arraystretch}{1} 
\begin{tabular}{cccccc}
\toprule
UCI1&&Mendeley&&UCI2\\
Feature&Class&Feature&Class&Feature&Class\\
\midrule
IPAddress	&1&	NumDots&1&	IPAddress&1	\\
\textbf{UrlLen}&1&	SubDomainLevl	&1&	\textbf{UrlLen}&1\\
ShortService	&1&	PathLevl	&1&	SSLfinalSt&1\\
AtSymbol	&1&	\textbf{UrlLen}	&1&	\textbf{ReqUrl}&2\\
\textbf{PrefSuff}&1& \textbf{NumDash}&1&	\textbf{UrlAnchor}&2\\
\textbf{HaveSubDomain}&1&	NumUnderscore&1&	\textbf{SFH}&2\\
SSLfinalSt	&1&	NumQueryComp&1&	 PopUpWin&3\\
DomainRegLen	&1&	NumNumcChars&1&	WebTraffic&4\\
\textbf{Favicon}&1& RandString&1&	AgeOfDomain&4\\
HTTPSToken&1	&	\textbf{DomainInPaths}&1&	$ $& $ $\\
\textbf{ReqUrl}&2	&	HostnameLen	&1&	$ $& $ $	\\
\textbf{UrlAnchor}&2	&	PathLen	&1&	$ $& $ $	\\
\textbf{LinksInTags}	&2&	DoubleSlashInPath&1&	$ $& $ $	\\
\textbf{SFH}&2	&	\textbf{ExtFavicon}	&1& $ $& $ $	\\
\textbf{Submit2Email}&2& 	\textbf{PctExtResUrls}&2&	$ $& $ $	\\
Redirect&3	&	\textbf{PctExtHlinks}&2&  $ $& $ $	\\
PopUpWin&3	&	InsecureForms	&2&	$ $& $ $	\\
AgeOfDomain&4	&	RelativeFormAction&2&	$ $& $ $	\\
DNSRecord&4	&	\textbf{PctNullSelfRedirHlinks}&2&	$ $& $ $	\\
WebTraffic&4	&	\textbf{Submit2Email}&2&	$ $& $ $	\\
PageRank	&4&	FreqDomainNameMismatch&3&	 $ $& $ $	\\
GoogleIndx&4	&	IframeOrFrame&3&	$ $& $ $	\\
LinksToPage&4	&	MissingTitle&3&	$ $& $ $	\\
StatisticalReport&4	&	\textbf{PctExtResUrlsRT}	&3&	$ $& $ $	\\
	$ $ & $ $	& \textbf{AbnormExtFormActR}	&3&	$ $& $ $	\\
	$ $ & $ $	&	\textbf{ExtMetaScriptLinkRT}	&3&	$ $& $ $	\\
	$ $ & $ $	&	\textbf{PctExtNullSelfRedirHlinksRT}	&3&	$ $& $ $	\\
		
\bottomrule
\end{tabular}
\begin{tablenotes}
      \small
      \item Classes 1, 2, 3, and 4 indicate Address bar-based, Abnormal-based, HTML/Javascript-based, and Domain-based feature classes, respectively.
\end{tablenotes}
\label{tab:features}
\vspace{-5px}
\end{threeparttable}
\end{table}

\section{Conclusion}
\label{sec:conclusion}

In this work, we consider finding an agreement about the definitive features which should be used in detecting phishing websites. We evaluate the detection performance using features selected by Fuzzy Rough Set (FRS) theory based on three benchmarked phishing data sets. After selecting the features, we utilize three well-known classification techniques and conduct the experiments by an unseen data set of 14,000 website samples to avoid classification overfitting. Comparisons with other feature selection methods utilized by the related work indicate the outperformance of FRS in selecting efficient features. The maximum F-measure gained by the features selected by FRS is 95\% using Random Forest classification. Also, there are 9 universal features selected by FRS over all the three data sets. The F-measure value using this universal feature set is approximately 93\% which is a comparable result in contrast to the FRS performance. Since the universal features contain no domain-based features, this finding implies that with no inquiry from external sources, we can gain a faster phishing detection which is also robust toward zero-day attacks.

\section{Acknowledgment}

This paper is based on work supported by the National Science Foundation (NSF) under Grant No. 1464104. Any opinions, findings, and conclusions or recommendations expressed are those of the author(s) and do not necessarily reflect the views of the NSF.

\bibliographystyle{IEEEtran}{}
\bibliography{IEEEfull}

\end{document}